\documentclass[letterpaper, 10 pt, conference]{ieeeconf}

\IEEEoverridecommandlockouts

\usepackage{graphicx, caption, booktabs, multirow, adjustbox, rotating, color, makecell}
\usepackage[hidelinks]{hyperref}
\usepackage[percent]{overpic}

\title{\LARGE \bf
Benchmarking Deep Learning Models for Aerial LiDAR Point Cloud Semantic Segmentation under Real Acquisition Conditions: A Case Study in Navarre
}

\author{Alex Salvatierra$^{1}$, José Antonio Sanz$^{1}$, Christian Gutiérrez$^{2}$, and Mikel Galar$^{1}$%
\thanks{$^{1}$Department of Statistics, Computer Science and Mathematics and Institute of Smart Cities (ISC), Public University of Navarre (UPNA), Campus de Arrosadía s/n, Pamplona, 31006, Navarre, Spain}%
\thanks{$^{2}$Tracasa Instrumental, Cabárceno, 6, Sarriguren, 31621, Navarre, Spain}%
}

\let\oldtwocolumn\twocolumn
\renewcommand\twocolumn[1][]{%
  \oldtwocolumn[{#1}{
    \vspace*{-1.5\baselineskip}
    \begin{center}
    \begin{overpic}[width=\textwidth]{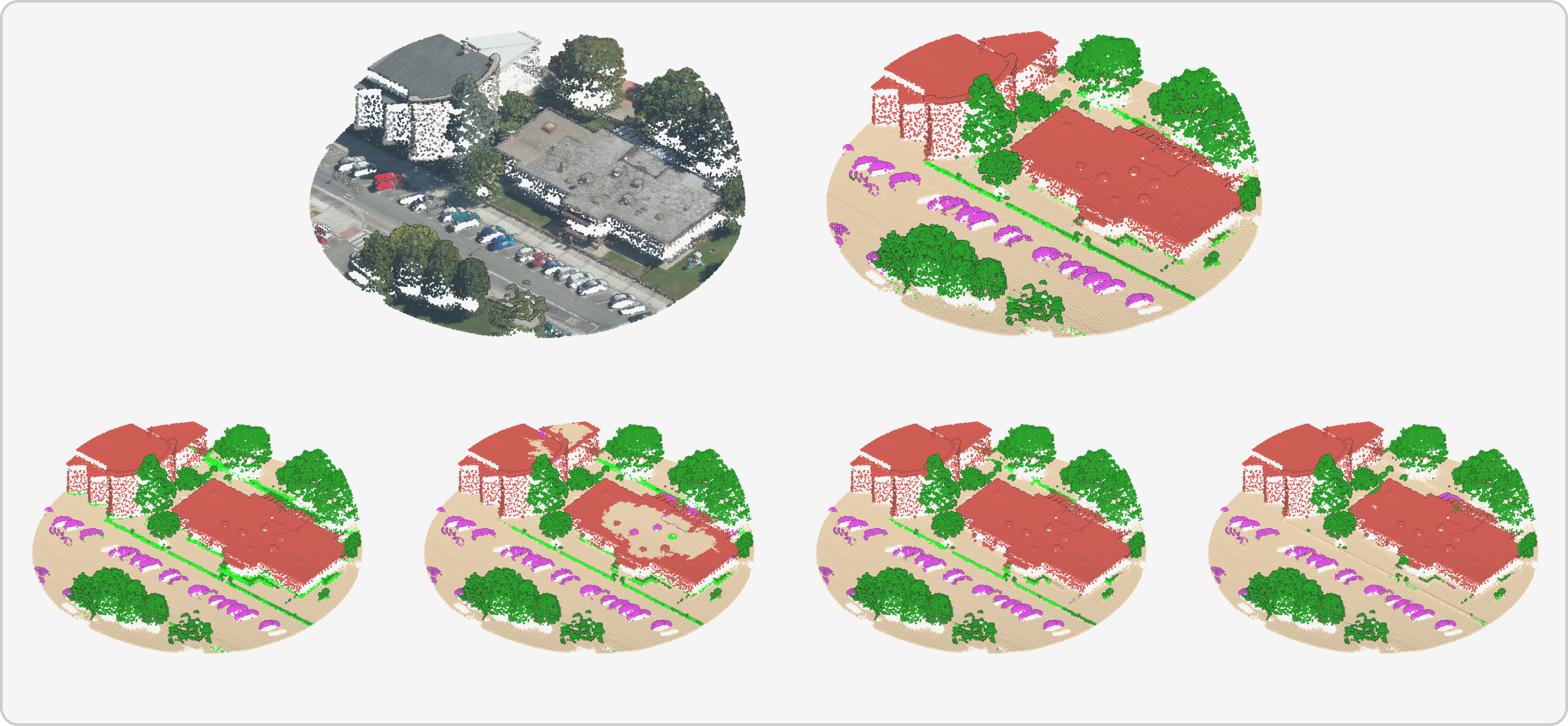}
                        
        \put(33.5, 22.5){\makebox(0,0){\textbf{RGB}}}
        \put(66.5, 22.5){\makebox(0,0){\textbf{Ground Truth}}}
        
        \put(12.5, 2.5){\makebox(0,0){\textbf{KPConv}}}
        \put(37.5, 2.5){\makebox(0,0){\textbf{RandLA-Net}}}
        \put(62.5, 2.5){\makebox(0,0){\textbf{SPT}}}
        \put(87.5, 2.5){\makebox(0,0){\textbf{PTv3}}}
    \end{overpic}
    
    \captionof{figure}{\textbf{Example scene segmented by four deep learning models.}
    Top row shows the input RGB scene and its corresponding ground truth semantic labels.
    Bottom row presents the same scene segmented by four models: \textit{KPConv}, \textit{RandLA-Net}, \textit{Superpoint Transformer} (SPT), and \textit{Point Transformer V3} (PTv3).
    Each point is colored according to its semantic class: ground (beige), low vegetation (bright green), medium/high vegetation (fern green), building (vermilion), and vehicle (magenta).}
    \label{fig:qualitative_comparison}
    \end{center}
  }]
}

\begin{document}

\setlength{\heavyrulewidth}{0.08em}
\setlength{\lightrulewidth}{0.05em}
\setlength{\cmidrulewidth}{0.03em}

\maketitle
\thispagestyle{empty}
\pagestyle{empty}

\begin{abstract}

Recent advances in deep learning have significantly improved 3D semantic segmentation, but most models focus on indoor or terrestrial datasets. Their behavior under real aerial acquisition conditions remains insufficiently explored, and although a few studies have addressed similar scenarios, they differ in dataset design, acquisition conditions, and model selection. To address this gap, we conduct an experimental benchmark evaluating several state-of-the-art architectures on a large-scale aerial LiDAR dataset acquired under operational flight conditions in Navarre, Spain, covering heterogeneous urban, rural, and industrial landscapes. This study compares four representative deep learning models, including KPConv, RandLA-Net, Superpoint Transformer, and Point Transformer V3, across five semantic classes commonly found in airborne surveys, such as ground, vegetation, buildings, and vehicles, highlighting the inherent challenges of class imbalance and geometric variability in aerial data. Results show that all tested models achieve high overall accuracy exceeding 93\%, with KPConv attaining the highest mean IoU (78.51\%) through consistent performance across classes, particularly on challenging and underrepresented categories. Point Transformer V3 demonstrates superior performance on the underrepresented vehicle class (75.11\% IoU), while Superpoint Transformer and RandLA-Net trade off segmentation robustness for computational efficiency.

\end{abstract}

\section{INTRODUCTION} \label{introduction}

Aerial Light Detection and Ranging (LiDAR) \cite{wangLiDARPointClouds2018} has become an essential technology for acquiring large-scale three-dimensional information about the Earth’s surface. Airborne laser scanning systems (ALS) \cite{wehrAirborneLaserScanning1999} enable the reconstruction of terrain, vegetation, and urban structures with high geometric accuracy, supporting applications in environmental monitoring \cite{wulderLidarSamplingLargearea2012}, land management \cite{yanUrbanLandCover2015}, and infrastructure assessment \cite{ferozUavbasedRemoteSensing2021}. As the spatial resolution and coverage of these datasets increase, automated semantic segmentation methods have gained critical importance for transforming raw point clouds into meaningful geospatial products, such as Digital Surface Models (DSMs) and Digital Terrain Models (DTMs) \cite{mooreDigitalTerrainModelling1991}.

Deep learning has driven major advances in 3D semantic segmentation during the last decade. Following the foundational PointNet \cite{charlesPointNetDeepLearning2017}, which pioneered direct processing of unordered point sets, subsequent architectures based on hierarchical sampling \cite{qiPointNetDeepHierarchical2017}, convolutions \cite{thomasKPConvFlexibleDeformable2019}, and graph-based approaches \cite{huRandLANetEfficientSemantic2020}, along with recent transformer-based models \cite{laiStratifiedTransformer3D2022, wuPointTransformerV22022, robertEfficient3DSemantic2023b, wuPointTransformerV32024}, have achieved increasingly strong performance on well-established benchmarks, particularly in indoor \cite{armeni3DSemanticParsing2016, daiScanNetRichlyAnnotated3D2017} and terrestrial \cite{behleySemanticKITTIDatasetSemantic2019, hackelSEMANTIC3DNETNEWLARGESCALE2017} domains. However, most of these datasets differ substantially from the characteristics of aerial LiDAR data, which typically exhibit irregular densities, strong class imbalance, and large variations in object scale. As a result, the transferability of existing 3D segmentation models to airborne scenarios remains only partially understood.

Compared to terrestrial or urban mapping settings, aerial LiDAR segmentation faces unique challenges. The nadir perspective and flight parameters lead to heterogeneous point distributions, while the large geographic coverage introduces high variability in scene composition. Classes such as vegetation, ground, and buildings dominate most scenes, whereas smaller objects like vehicles are underrepresented. These factors make large-scale evaluation essential for assessing model robustness under realistic acquisition conditions.

Recent benchmarks such as DALES \cite{varneyDALESLargescaleAerial2020} and FRACTAL \cite{gaydonFRACTALUltraLargeScaleAerial2024} have broadened the geographic and semantic diversity of aerial LiDAR data, advancing the application of deep learning to airborne segmentation tasks. However, most publicly available datasets, including DALES and FRACTAL, remain limited in some of the following aspects: spatial extent, point density, scope of evaluated methods, or the diversity of environments represented. Moreover, the emergence of newer architectures, particularly transformer-based models, underscores the need for updated comparative evaluations on aerial data, as these approaches have remained largely unevaluated in this domain.

To address this gap, the objective of this contribution is to benchmark four representative methods: KPConv \cite{thomasKPConvFlexibleDeformable2019}, RandLA-Net \cite{huRandLANetEfficientSemantic2020}, Superpoint Transformer \cite{robertEfficient3DSemantic2023b}, and Point Transformer V3 \cite{wuPointTransformerV32024}, spanning convolutional, MLP-based, and transformer architectures, on operational aerial LiDAR data acquired in Navarre, Spain, across heterogeneous urban, rural, and industrial landscapes. The evaluation incorporates recent transformer-based models that represent the current state-of-the-art but remain largely unevaluated on operational aerial data, assessing their performance across unfiltered scenes with natural class imbalance and variable point densities. Consequently, this study is aimed at assessing model robustness, class-level performance, and generalization capabilities under these conditions, complementing existing benchmarks with insights into the current state of deep learning performance on aerial LiDAR data.

The remainder of the paper is organized as follows. Section~\ref{related} reviews related work. Section~\ref{setup} describes the experimental setup, including data, models, training protocol, and evaluation metrics. Section~\ref{results} presents the quantitative and qualitative results along with comparative analysis. Section~\ref{conclusions} concludes the paper.

\section{RELATED WORK} \label{related}

Semantic segmentation of 3D point clouds has advanced rapidly with the development of deep learning architectures. 
Early methods such as PointNet \cite{charlesPointNetDeepLearning2017} established the foundations for directly processing unordered point sets, while PointNet++ \cite{qiPointNetDeepHierarchical2017} introduced hierarchical multi-scale aggregation. 
Subsequent approaches explored different paradigms: convolution-based methods including KPConv \cite{thomasKPConvFlexibleDeformable2019}, PointCNN \cite{liPointCNNConvolutionXTransformed2018}, and PointConv \cite{wuPointConvDeepConvolutional2019}; efficient MLP-based architectures such as RandLA-Net \cite{huRandLANetEfficientSemantic2020}; and graph-based approaches like DGCNN \cite{wangDynamicGraphCNN2019} and Superpoint Graphs \cite{landrieuLargeScalePointCloud2018}. 
More recently, transformer-based models \cite{guoPCTPointCloud2021, robertEfficient3DSemantic2023b, zhaoPointTransformer2021, wuPointTransformerV22022, wuPointTransformerV32024} have achieved state-of-the-art performance by modeling point sets through attention mechanisms.

However, most of these architectures were developed and benchmarked on indoor or terrestrial datasets such as S3DIS \cite{armeni3DSemanticParsing2016}, ScanNet \cite{daiScanNetRichlyAnnotated3D2017}, SemanticKITTI \cite{behleySemanticKITTIDatasetSemantic2019}, and Semantic3D \cite{hackelSEMANTIC3DNETNEWLARGESCALE2017}. These datasets differ markedly from airborne LiDAR in acquisition perspective, sampling density, spatial extent, and class distribution, making the direct transferability of conclusions to aerial contexts uncertain.

Deep learning for airborne LiDAR has gained attention more recently with dedicated ALS benchmarks. The ISPRS Vaihingen benchmark \cite{rottensteinerISPRSBENCHMARKURBAN2012} was the first to standardize semantic labeling for aerial LiDAR with nine classes, though its limited coverage (0.2~km²) and low point density (4–7~pts/m²) make it inadequate for training modern deep learning architectures.
Early large-scale efforts such as DFC 2018 \cite{xuAdvancedMultiSensorOptical2019} and DublinCity \cite{zolanvariDublinCityAnnotatedLiDAR2019} expanded spatial coverage and point density respectively, but remained limited in environmental diversity or geographic extent.

DALES \cite{varneyDALESLargescaleAerial2020} represented a major milestone, covering 10~km² with 500 million labeled points at 50~pts/m² and eight classes, evaluating multiple architectures including KPConv, PointNet++, and Superpoint Graphs on urban scenes. Subsequent contributions addressed specific aspects: LASDU \cite{yeLASDULargeScaleAerial2020} focused on dense urban environments, OpenGF \cite{qinOpenGFUltraLargeScaleGround2021} provided the largest publicly available dataset (47~km²) for ground filtering, and YUTO Semantic \cite{yooYUTOSEMANTICLARGE2023} performed cross-model evaluations (KPConv, RandLA-Net, EyeNet \cite{yooEyeNetMultiscaleMultidensity2025}) across multiple flight missions, albeit limited to urban environments.
Additional specialized datasets include CENAGIS-ALS \cite{zacharCENAGISALSBENCHMARKNEW2023}, notable for its extremely high density (275~pts/m²) over urban blocks.

Most recently, FRACTAL \cite{gaydonFRACTALUltraLargeScaleAerial2024} introduced an ultra-large-scale dataset covering 250~km² of heterogeneous French landscapes with 9.26 billion labeled points at 37~pts/m², offering unprecedented geographic diversity spanning rural, urban, and forested environments, though evaluating only a RandLA-Net baseline. ECLAIR \cite{melekhovECLAIRHighFidelityAerial2024} provided substantial coverage (10~km²) with high point density (50~pts/m²) and eleven semantic classes, though evaluating only Minkowski-based \cite{choy4DSpatioTemporalConvNets2019} architectures.

Despite these advances, most publicly available ALS datasets remain limited in spatial extent, point density, evaluated architectural diversity, or environmental representation. Many rely on carefully controlled acquisition or urban-centric coverage, and the rapid emergence of transformer-based architectures has outpaced benchmark development, with recent transformer-based models such as Superpoint Transformer \cite{robertEfficient3DSemantic2023b} and Point Transformer~V3 \cite{wuPointTransformerV32024} largely unevaluated on aerial data.

In response, this study benchmarks four representative architectures: KPConv, RandLA-Net, Superpoint Transformer, and Point Transformer V3, spanning convolutional, MLP-based, and transformer paradigms, on operationally acquired aerial LiDAR across heterogeneous landscapes, providing updated comparative insights into model robustness and class-level performance under realistic conditions.

\section{EXPERIMENTAL SETUP} \label{setup}

\subsection{Data and Preprocessing} \label{data}

\begin{figure}[t]
    \centering
    \includegraphics[width=\columnwidth]{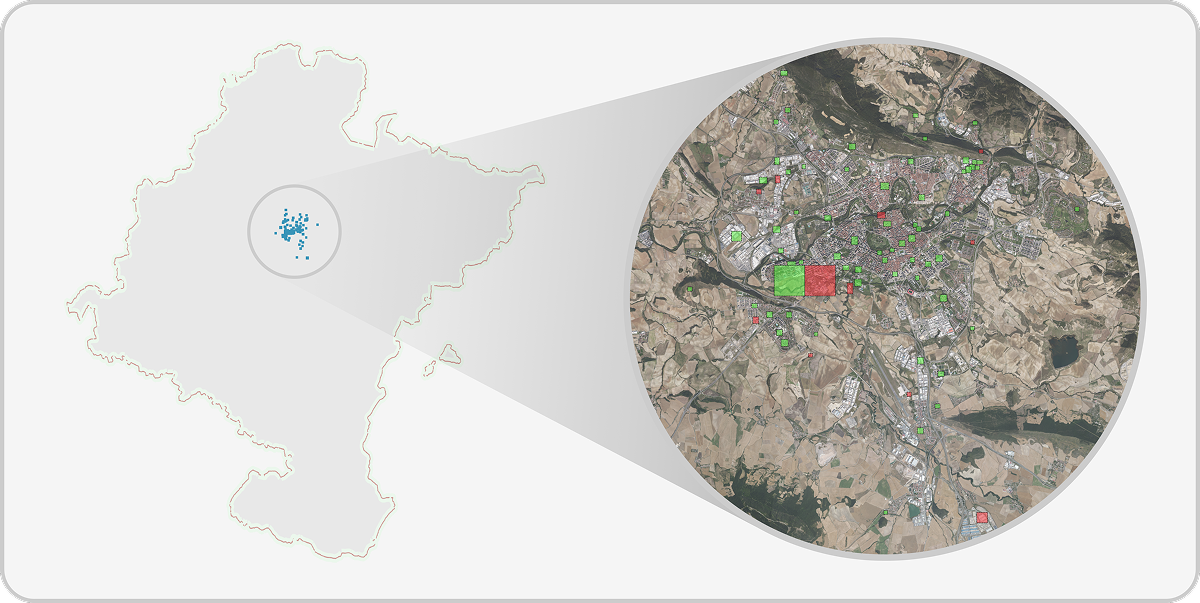}
    \caption{Geographical distribution of the evaluated LiDAR scenes around Pamplona, Navarre, Spain. {\color{green}Green} areas indicate training zones, while {\color{red}red} areas represent the test set.}
    \label{fig:dataset_map}
\end{figure}

The experiments were conducted on a large-scale aerial LiDAR dataset acquired under operational flight conditions in the Pamplona area (Navarre, Spain). The survey covers approximately 4~km² encompassing urban, industrial, rural, and semi-rural environments, as illustrated in Fig.~\ref{fig:dataset_map}, split into training (65\%) and test (35\%) sets, with an average point density of about 50~pts/m². Each point is described by its three-dimensional coordinates $(x,y,z)$, intensity, RGB color, return number, number of returns, and normalized difference vegetation index (NDVI).

The dataset comprises five semantic classes representative of typical airborne mapping scenarios: \textit{ground}, \textit{low vegetation}, \textit{medium/high vegetation}, \textit{building}, and \textit{vehicle}. Table~\ref{tab:class_distribution} summarizes the point distribution per semantic class, highlighting significant class imbalance characteristic of operational aerial data, especially for the \textit{vehicle} and \textit{low vegetation} categories, which together account for around 2\% of the labeled points.

For training and evaluation, the original point clouds were partitioned into overlapping $50 \times 50$~m tiles, and both coordinates and per-point attributes were normalized prior to training. At inference time, predictions from overlapping tiles were merged by averaging class probabilities at point level, providing additional spatial context and implicit test-time augmentation.

\begin{table}[t]
\centering
\caption{Labeled points per class in millions (M) and percentage (\%).}\label{tab:class_distribution}
\setlength{\tabcolsep}{6pt}
\begin{tabular}{lrr}
\toprule
\textbf{Class} & \textbf{Points (M)} & \textbf{\%} \\
\midrule
Ground              & 87.99  & 62.50 \\
Low Vegetation      & 1.98   & 1.41 \\
Medium/High Veg.    & 22.77  & 16.17 \\
Building            & 27.09  & 19.24 \\
Vehicle             & 0.96   & 0.68 \\
\midrule
\textbf{Total}      & \textbf{140.80} & \textbf{100} \\
\bottomrule
\end{tabular}
\end{table}

\subsection{Models} \label{models}

Four representative deep learning architectures were selected for comparison: KPConv, RandLA-Net, Superpoint Transformer (SPT), and Point Transformer V3. These models were chosen because they capture complementary paradigms in 3D point cloud processing, remain among the most widely adopted baselines in recent literature, and have consistently shown strong performance in public benchmarks.

\noindent\textbf{KPConv} \cite{thomasKPConvFlexibleDeformable2019} is a convolution-based architecture that defines continuous convolution kernels on point neighborhoods through kernel point sets. It has demonstrated high accuracy across diverse domains due to its ability to preserve fine spatial detail and robustly model local surfaces.

\noindent\textbf{RandLA-Net} \cite{huRandLANetEfficientSemantic2020} represents a lightweight and scalable approach specifically designed for large-scale point clouds. It combines random sampling with local feature aggregation layers, significantly reducing memory consumption. This makes it particularly suitable for dense aerial LiDAR data where millions of points must be processed per scene.

\noindent\textbf{Superpoint Transformer (SPT)} \cite{robertEfficient3DSemantic2023b} introduces an attention-based framework that operates on geometrically coherent superpoints instead of individual points. By combining spatial grouping with self-attention, it efficiently captures long-range contextual relationships while preserving geometric structure.

\noindent\textbf{Point Transformer V3 (PTv3)} \cite{wuPointTransformerV32024} abandons the strict permutation-invariant treatment of point sets by serializing point clouds into ordered sequences and applying patch-wise attention. This replaces costly KNN-based neighborhoods and heavy positional encodings with a simpler, more efficient scheme that greatly expands the receptive field.

All models were implemented using publicly available source code\footnote{
KPConv: \url{https://github.com/torch-points3d/torch-points3d};
RandLA-Net: \url{https://github.com/IGNF/myria3d};
SPT: \url{https://github.com/drprojects/superpoint_transformer};
PTv3: \url{https://github.com/Pointcept/Pointcept}.} and trained under the same experimental conditions to ensure comparability. Their configurations followed recommended hyperparameter settings from the original publications, with minor adjustments to accommodate the spatial scale and density of aerial LiDAR data.

\begin{table*}[t]
\centering
\caption{Overview of the selected methods on the dataset test set. We report the overall accuracy, mean IoU, and per-class IoU for each category. The best values in each column are highlighted in \textbf{bold}.}
\label{tab:oa_iou_models}
\begin{tabular}{l c c c c c c c}
\toprule
\multirow{2}{*}{\raisebox{-3.8ex}{Method}}
& \multirow{2}{*}{\raisebox{-3.8ex}{OA}}
& \multicolumn{6}{c}{IoU} \\ 
\cmidrule(lr){3-8}
 &  & \textit{Mean} & \textit{Ground} & \textit{Low Veg.} & \textit{Med./High Veg.} & \textit{Building} & \textit{Vehicle} \\
\midrule
KPConv \cite{thomasKPConvFlexibleDeformable2019} & \textbf{96.16} & \textbf{78.51} & \textbf{95.11} & \textbf{33.61} & \textbf{95.12} & \textbf{93.97} & 74.76 \\
RandLA-Net \cite{huRandLANetEfficientSemantic2020} & 93.39 & 71.98 & 91.23 & 23.61 & 93.57 & 82.49 & 69.02 \\
SPT \cite{robertEfficient3DSemantic2023b} & 95.37 & 74.27 & 94.42 & 20.61 & 91.20 & 92.23 & 72.87 \\
PTv3 \cite{wuPointTransformerV32024} & 95.79 & 73.56 & 94.68 & 11.23 & 93.10 & 93.69 & \textbf{75.11} \\
\bottomrule
\end{tabular}
\end{table*}

\subsection{Training Protocol} \label{protocol}

All models were trained using the same experimental setup to ensure fair comparison. Training was conducted on a workstation equipped with an NVIDIA RTX 6000 Ada GPU with 48 GB of VRAM, an AMD EPYC 9454 processor, and 64 GB of RAM. Each model was trained three times with different random seeds, and the reported results correspond to the average performance across these runs to account for training variability.

All models used a grid size of 0.005 in the normalized coordinate space, resulting in an effective sampling density of approximately 20 points per square meter. Batch sizes were adjusted according to each architecture's requirements: 24 for RandLA-Net, 10 for KPConv and Point Transformer V3, and 4 for Superpoint Transformer. The dataset was partitioned into training and test sets, and models were trained until convergence. 

\subsection{Evaluation Metrics} \label{metrics}

Model performance was assessed using standard metrics for semantic segmentation of 3D point clouds, including Overall Accuracy (OA), per-class Intersection-over-Union (IoU) and mean IoU (mIoU). These metrics jointly quantify both global accuracy and class-wise consistency, providing a comprehensive evaluation across dominant and minority categories.

Per-class IoU (Eq.~\ref{eq:iou}), mean IoU (Eq.~\ref{eq:miou}), and Overall Accuracy (Eq.~\ref{eq:oa}) are defined as:

\begin{equation}
\mathrm{IoU}_c = \frac{TP_c}{TP_c + FP_c + FN_c}
\label{eq:iou}
\end{equation}

\begin{equation}
\mathrm{mIoU} = \frac{1}{N_c} \sum_{c=1}^{N_c} \mathrm{IoU}_c
\label{eq:miou}
\end{equation}

\begin{equation}
\mathrm{OA} = \frac{\sum_{c=1}^{N_c} TP_c}{\sum_{c=1}^{N_c} (TP_c + FN_c)}
\label{eq:oa}
\end{equation}

where $TP_c$, $FP_c$, and $FN_c$ denote the number of true positive, false positive, and false negative points for class $c$, respectively, and $N_c$ is the total number of classes.

While OA summarizes overall correctness, mIoU offers more informative insights under class imbalance by weighting all categories equally. Reporting both global and per-class metrics ensures a fair comparison among models, particularly for underrepresented classes.

\begin{table}[t]
\centering
\caption{Computational efficiency comparison. Parameters are reported in millions (M) and times in minutes (min). Training corresponds to the total duration until convergence, and inference to the processing of the full test set.}
\label{tab:efficiency}
\resizebox{\columnwidth}{!}{%
    \begin{tabular}{l c c c c}
    \toprule
    \textbf{Model} & \textbf{\makecell{Params \\ (M)}} & \textbf{Epochs} & \textbf{\makecell{Training \\ (min)}} & \textbf{\makecell{Inference \\ (min)}} \\
    \midrule
    KPConv \cite{thomasKPConvFlexibleDeformable2019} & 23.26 & 20  & 611 & 40 \\
    RandLA-Net \cite{huRandLANetEfficientSemantic2020} & 1.11  & 80  & 307 & 17 \\
    SPT \cite{robertEfficient3DSemantic2023b} & 0.21  & 80  & 513 & 19 \\
    PTv3 \cite{wuPointTransformerV32024} & 46.18 & 120 & 531 & 14 \\
    \bottomrule
    \end{tabular}%
}
\end{table}

\section{RESULTS} \label{results}

This section presents the quantitative and qualitative results of the benchmark. The analysis focuses on both global metrics and per-class performance to identify differences in model behavior across dominant and minority categories, as well as their computational efficiency.

\subsection{Quantitative Performance} \label{quantitative}

Table~\ref{tab:oa_iou_models} summarizes the overall performance of the four evaluated architectures on the dataset test set. The results reveal that all models achieve high overall accuracy, ranging from 93.39\% (RandLA-Net) to 96.16\% (KPConv), demonstrating the general capability of modern deep learning approaches for aerial LiDAR segmentation. However, the mean IoU metric, which weights all classes equally, shows more pronounced differences, spanning from 71.98\% to 78.51\%, indicating varying degrees of robustness across semantic categories. Table~\ref{tab:efficiency} presents the computational cost comparison, revealing significant trade-offs between model complexity, training duration, and inference speed.

KPConv emerges as the best-performing model, achieving the highest overall accuracy (96.16\%) and mean IoU (78.51\%). Remarkably, it leads in four out of five semantic classes, including \textit{ground} (95.11\%), \textit{low vegetation} (33.61\%), \textit{medium/high vegetation} (95.12\%), and \textit{building} (93.97\%). However, this superior precision comes at a significant computational cost. As shown in Table~\ref{tab:efficiency}, KPConv requires the longest training (611 min over 20 epochs) and inference (40 min) times with 23.26~M parameters, reflecting its heavy computational load.

RandLA-Net, despite achieving the lowest mean IoU (71.98\%), maintains competitive performance on dominant classes such as \textit{ground} (91.23\%) and \textit{medium/high vegetation} (93.57\%). The random sampling strategy, while computationally efficient, appears to lose critical geometric details necessary for accurate classification of minority and boundary regions, as evidenced by its performance on \textit{low vegetation} (23.61\%) and \textit{building} (82.49\%). With only 1.11~M parameters, it demonstrates the fastest training time (307 minutes over 80 epochs) and second-fastest inference (17 minutes), highlighting its computational efficiency.

Superpoint Transformer demonstrates balanced performance across most classes, ranking second in mean IoU (74.27\%). Its superpoint-based aggregation provides a good trade-off between computational efficiency and segmentation quality. However, it underperforms on \textit{low vegetation} (20.61\%), indicating that geometric grouping at the superpoint level may merge sparse vegetation with surrounding terrain. With only 0.21M parameters, SPT is the most lightweight architecture.

Point Transformer V3 achieves the highest performance on the \textit{vehicle} class (75.11\% IoU), surpassing KPConv by 0.35 points, validating attention mechanisms for small, isolated objects despite their scarcity in the training data. However, PTv3 shows the lowest performance on \textit{low vegetation} (11.23\% IoU), suggesting that the serialization-based grouping strategy may struggle with diffuse, low-density point patterns. Despite having the highest parameter count (46.18M), it achieves the fastest inference time (14 minutes) due to its optimized architecture.

The most striking observation is the performance gap between majority and minority classes across all models. While all architectures achieve IoU above 91\% on \textit{ground}, \textit{medium/high vegetation}, and \textit{building}, the results on \textit{low vegetation} range from 11.23\% to 33.61\%. This reflects the severe class imbalance present in operational aerial LiDAR data, where \textit{low vegetation} represents only 1.41\% of the total points (see Table~\ref{tab:class_distribution}). The \textit{vehicle} class, though similarly underrepresented (0.68\%), achieves notably higher IoU across all models (69.02\%–75.11\%), likely due to its more distinctive geometric structure compared to the diffuse nature of low vegetation.

\subsection{Qualitative Analysis} \label{qualitative}

Figure~\ref{fig:qualitative_comparison} shows a representative test scene segmented by the four evaluated architectures, alongside the RGB image and ground truth labels. All models successfully identify the main structural components yet differ markedly in boundary precision, minority-class consistency, and treatment of ambiguous regions.

Among the semantic classes, \textit{low vegetation} is consistently the most challenging. Its geometry closely overlaps with both \textit{ground} and \textit{medium/high vegetation}, forming thin, discontinuous layers. This makes it far easier to confuse than compact and geometrically distinctive objects such as \textit{vehicles}, despite also being highly underrepresented.

KPConv delivers the most coherent segmentation. Building roofs and vegetation boundaries are sharply delineated with minimal bleeding, and recovers the highest volume of \textit{low vegetation}, demonstrating superior sensitivity to this challenging class compared to other architectures.

RandLA-Net exhibits the most evident errors. Large planar roofs are frequently misclassified as \textit{ground}, and ambiguous structures also collapse into this class. Small \textit{vehicle} predictions occasionally appear on top of buildings or flat surfaces, reflecting the boundary ambiguity introduced by its random sampling strategy.

Superpoint Transformer produces clean and stable \textit{building} masks. Visual inspection reveals robust segmentation performance even in \textit{low vegetation} regions. Notably, it correctly identifies a complex building structure that proves challenging for other architectures, demonstrating its robustness in handling complex geometries.

Point Transformer V3 achieves the most reliable \textit{vehicle} detection, benefiting from its wide receptive field and attention-based feature weighting. However, it shows mild bleeding between \textit{vegetation} and \textit{building} in areas where both classes are spatially intertwined, and it rarely predicts \textit{low vegetation}, which is often absorbed into the \textit{ground} class.

Overall, the qualitative inspection reinforces the quantitative results, illustrating how performance variations translate into observable differences in boundary precision, minority-class recovery, and geometric detail preservation.

\section{CONCLUSIONS} \label{conclusions}

This work benchmarks four state-of-the-art deep learning architectures: KPConv, RandLA-Net, Superpoint Transformer, and Point Transformer V3, on operational aerial LiDAR point clouds covering heterogeneous landscapes in Navarre, Spain. The experiments reflect practical airborne mapping scenarios characterized by strong class imbalance.

Results show that all architectures achieve high overall accuracy ($>$93\%), but mean IoU shows more pronounced differences (71.98\%–78.51\%). This disparity arises because overall accuracy is dominated by majority classes, while mean IoU weights all categories equally, exposing difficulties with minority classes. In this context, KPConv stands out as the best-performing model overall, showing robustness across all semantic classes and particularly excelling on minority categories such as \textit{low vegetation} (33.61\% IoU). Point Transformer V3 performs well on minority classes with clear geometric structures, particularly \textit{vehicles} (75.11\% IoU). Superpoint Transformer achieves competitive results with high parameter efficiency, whereas RandLA-Net exhibits lower overall performance as a trade-off for its computational efficiency.

The persistent difficulties with \textit{low vegetation}, where IoU ranges from 11.23\% to 33.61\%, highlight the inherent geometric ambiguity of near-ground vegetation layers and the challenges posed by severe class imbalance under real flight conditions. This underscores the need for specialized strategies to improve minority class recognition in operational aerial LiDAR segmentation.

This benchmark provides quantitative and qualitative evidence of current model capabilities and limitations under realistic conditions, contributing to bridging the gap between controlled research environments and practical large-scale deployment. Future work should explore evaluation methodologies that go beyond point-wise metrics to better capture spatial consistency, as well as architectural or training strategies explicitly tailored to aerial LiDAR characteristics.

\section*{ACKNOWLEDGMENT}

Alex Salvatierra holds a predoctoral scholarship funded by the Tracasa Chair in Computer Science and Artificial Intelligence at the Public University of Navarre. This work was also supported by project PID2022-136627NB-I00 (MCIN/AEI/10.13039/501100011033/FEDER, EU).

\bibliographystyle{IEEEtran}
\bibliography{IEEE_CAI}

@inproceedings{armeni3DSemanticParsing2016,
  title = {{{3D Semantic Parsing}} of {{Large-Scale Indoor Spaces}}},
  booktitle = {Proceedings of the {{IEEE Conference}} on {{Computer Vision}} and {{Pattern Recognition}}},
  author = {Armeni, Iro and Sener, Ozan and Zamir, Amir R. and Jiang, Helen and Brilakis, Ioannis and Fischer, Martin and Savarese, Silvio},
  year = 2016,
  pages = {1534--1543},
  urldate = {2025-10-21}
}

@article{behleySemanticKITTIDatasetSemantic2019,
  title = {{{SemanticKITTI}}: {{A Dataset}} for {{Semantic Scene Understanding}} of {{LiDAR Sequences}}},
  author = {Behley, Jens and Garbade, Martin and Milioto, Andres and Quenzel, Jan and Behnke, Sven and Stachniss, Cyrill and Gall, Jurgen},
  year = 2019,
  journal = {2019 IEEECVF Int. Conf. Comput. Vis. ICCV},
  pages = {9296--9306},
  publisher = {IEEE},
  address = {Seoul, Korea (South)},
  doi = {10.1109/ICCV.2019.00939},
  urldate = {2025-10-21},
  copyright = {https://ieeexplore.ieee.org/Xplorehelp/downloads/license-information/IEEE.html}
}

@inproceedings{charlesPointNetDeepLearning2017,
  title = {{{PointNet}}: {{Deep Learning}} on {{Point Sets}} for {{3D Classification}} and {{Segmentation}}},
  booktitle = {2017 {{IEEE Conf}}. {{Comput}}. {{Vis}}. {{Pattern Recognit}}. {{CVPR}}},
  author = {Charles, R. Qi and Su, Hao and Kaichun, Mo and Guibas, Leonidas J.},
  year = 2017,
  pages = {77--85},
  doi = {10.1109/CVPR.2017.16},
  urldate = {2025-09-01}
}

@inproceedings{choy4DSpatioTemporalConvNets2019,
  title = {{{4D Spatio-Temporal ConvNets}}: {{Minkowski Convolutional Neural Networks}}},
  booktitle = {2019 {{IEEECVF Conf}}. {{Comput}}. {{Vis}}. {{Pattern Recognit}}. {{CVPR}}},
  author = {Choy, Christopher and Gwak, JunYoung and Savarese, Silvio},
  year = 2019,
  pages = {3070--3079},
  doi = {10.1109/CVPR.2019.00319},
  urldate = {2025-09-01}
}

@inproceedings{daiScanNetRichlyAnnotated3D2017,
  title = {{{ScanNet}}: {{Richly-Annotated 3D Reconstructions}} of {{Indoor Scenes}}},
  booktitle = {Proceedings of the {{IEEE Conference}} on {{Computer Vision}} and {{Pattern Recognition}}},
  author = {Dai, Angela and Chang, Angel X. and Savva, Manolis and Halber, Maciej and Funkhouser, Thomas and Niessner, Matthias},
  year = 2017,
  pages = {5828--5839},
  urldate = {2025-10-21}
}

@article{ferozUavbasedRemoteSensing2021,
  title = {Uav-Based Remote Sensing Applications for Bridge Condition Assessment},
  author = {Feroz, Sainab and Dabous, Saleh Abu},
  year = 2021,
  journal = {Remote Sens.},
  volume = {13},
  number = {9},
  publisher = {MDPI AG},
  doi = {10.3390/rs13091809},
  langid = {english}
}

@misc{gaydonFRACTALUltraLargeScaleAerial2024,
  title = {{{FRACTAL}}: {{An Ultra-Large-Scale Aerial Lidar Dataset}} for {{3D Semantic Segmentation}} of {{Diverse Landscapes}}},
  author = {Gaydon, Charles and Daab, Michel and Roche, Floryne},
  year = 2024,
  number = {arXiv:2405.04634},
  eprint = {2405.04634},
  primaryclass = {cs},
  publisher = {arXiv},
  doi = {10.48550/arXiv.2405.04634},
  urldate = {2025-09-01},
  archiveprefix = {arXiv}
}

@article{guoPCTPointCloud2021,
  title = {{{PCT}}: {{Point}} Cloud Transformer},
  author = {Guo, Meng-Hao and Cai, Jun-Xiong and Liu, Zheng-Ning and Mu, Tai-Jiang and Martin, Ralph R. and Hu, Shi-Min},
  year = 2021,
  journal = {Comp. Visual Media},
  volume = {7},
  number = {2},
  pages = {187--199},
  doi = {10.1007/s41095-021-0229-5},
  urldate = {2025-09-01},
  langid = {english}
}

@article{hackelSEMANTIC3DNETNEWLARGESCALE2017,
  title = {{{SEMANTIC3D}}.{{NET}}: {{A NEW LARGE-SCALE POINT CLOUD CLASSIFICATION BENCHMARK}}},
  author = {Hackel, T. and Savinov, N. and Ladicky, L. and Wegner, J. D. and Schindler, K. and Pollefeys, M.},
  year = 2017,
  journal = {ISPRS Ann. Photogramm. Remote Sens. Spat. Inf. Sci.},
  volume = {IV-1-W1},
  pages = {91--98},
  publisher = {Copernicus GmbH},
  doi = {10.5194/isprs-annals-IV-1-W1-91-2017},
  urldate = {2025-10-21},
  langid = {english}
}

@inproceedings{huRandLANetEfficientSemantic2020,
  title = {{{RandLA-Net}}: {{Efficient Semantic Segmentation}} of {{Large-Scale Point Clouds}}},
  booktitle = {2020 {{IEEECVF Conf}}. {{Comput}}. {{Vis}}. {{Pattern Recognit}}. {{CVPR}}},
  author = {Hu, Qingyong and Yang, Bo and Xie, Linhai and Rosa, Stefano and Guo, Yulan and Wang, Zhihua and Trigoni, Niki and Markham, Andrew},
  year = 2020,
  pages = {11105--11114},
  doi = {10.1109/CVPR42600.2020.01112},
  urldate = {2025-09-01}
}

@inproceedings{laiStratifiedTransformer3D2022,
  title = {Stratified {{Transformer}} for {{3D Point Cloud Segmentation}}},
  booktitle = {2022 {{IEEECVF Conf}}. {{Comput}}. {{Vis}}. {{Pattern Recognit}}. {{CVPR}}},
  author = {Lai, Xin and Liu, Jianhui and Jiang, Li and Wang, Liwei and Zhao, Hengshuang and Liu, Shu and Qi, Xiaojuan and Jia, Jiaya},
  year = 2022,
  pages = {8490--8499},
  doi = {10.1109/CVPR52688.2022.00831},
  urldate = {2025-09-01}
}

@inproceedings{landrieuLargeScalePointCloud2018,
  title = {Large-{{Scale Point Cloud Semantic Segmentation}} with {{Superpoint Graphs}}},
  booktitle = {2018 {{IEEECVF Conf}}. {{Comput}}. {{Vis}}. {{Pattern Recognit}}.},
  author = {Landrieu, Loic and Simonovsky, Martin},
  year = 2018,
  pages = {4558--4567},
  doi = {10.1109/CVPR.2018.00479},
  urldate = {2025-09-01}
}

@inproceedings{liPointCNNConvolutionXTransformed2018,
  title = {{{PointCNN}}: {{Convolution On X-Transformed Points}}},
  booktitle = {Adv. {{Neural Inf}}. {{Process}}. {{Syst}}.},
  author = {Li, Yangyan and Bu, Rui and Sun, Mingchao and Wu, Wei and Di, Xinhan and Chen, Baoquan},
  year = 2018,
  volume = {31},
  publisher = {Curran Associates, Inc.},
  doi = {10.48550/arXiv.1801.07791},
  urldate = {2025-09-01}
}

@misc{melekhovECLAIRHighFidelityAerial2024,
  title = {{{ECLAIR}}: {{A High-Fidelity Aerial LiDAR Dataset}} for {{Semantic Segmentation}}},
  author = {Melekhov, Iaroslav and Umashankar, Anand and Kim, Hyeong-Jin and Serkov, Vladislav and Argyle, Dusty},
  year = 2024,
  number = {arXiv:2404.10699},
  eprint = {2404.10699},
  primaryclass = {cs},
  publisher = {arXiv},
  doi = {10.48550/arXiv.2404.10699},
  urldate = {2025-09-01},
  archiveprefix = {arXiv}
}

@article{mooreDigitalTerrainModelling1991,
  title = {Digital Terrain Modelling: {{A}} Review of Hydrological, Geomorphological, and Biological Applications},
  author = {Moore, I.D. and Grayson, R.B. and Ladson, A.R.},
  year = 1991,
  journal = {Hydrol. Processes},
  volume = {5},
  number = {1},
  pages = {3--30},
  doi = {10.1002/hyp.3360050103},
  langid = {english}
}

@misc{qinOpenGFUltraLargeScaleGround2021,
  title = {{{OpenGF}}: {{An Ultra-Large-Scale Ground Filtering Dataset Built Upon Open ALS Point Clouds Around}} the {{World}}},
  author = {Qin, Nannan and Tan, Weikai and Ma, Lingfei and Zhang, Dedong and Li, Jonathan},
  year = 2021,
  number = {arXiv:2101.09641},
  eprint = {2101.09641},
  primaryclass = {cs},
  publisher = {arXiv},
  doi = {10.48550/arXiv.2101.09641},
  urldate = {2025-09-01},
  archiveprefix = {arXiv}
}

@inproceedings{qiPointNetDeepHierarchical2017,
  title = {{{PointNet}}++: {{Deep Hierarchical Feature Learning}} on {{Point Sets}} in a {{Metric Space}}},
  booktitle = {Adv. {{Neural Inf}}. {{Process}}. {{Syst}}.},
  author = {Qi, Charles Ruizhongtai and Yi, Li and Su, Hao and Guibas, Leonidas J},
  year = 2017,
  volume = {30},
  publisher = {Curran Associates, Inc.},
  doi = {10.48550/arXiv.1706.02413},
  urldate = {2025-09-01}
}

@inproceedings{robertEfficient3DSemantic2023b,
  title = {Efficient {{3D Semantic Segmentation}} with {{Superpoint Transformer}}},
  booktitle = {2023 {{IEEECVF Int}}. {{Conf}}. {{Comput}}. {{Vis}}. {{ICCV}}},
  author = {Robert, Damien and Raguet, Hugo and Landrieu, Loic},
  year = 2023,
  pages = {17149--17158},
  doi = {10.1109/ICCV51070.2023.01577},
  urldate = {2025-09-30}
}

@article{rottensteinerISPRSBENCHMARKURBAN2012,
  title = {{{THE ISPRS BENCHMARK ON URBAN OBJECT CLASSIFICATION AND 3D BUILDING RECONSTRUCTION}}},
  author = {Rottensteiner, F. and Sohn, G. and Jung, J. and Gerke, M. and Baillard, C. and Benitez, S. and Breitkopf, U.},
  year = 2012,
  journal = {ISPRS Ann. Photogramm. Remote Sens. Spat. Inf. Sci.},
  volume = {I-3},
  pages = {293--298},
  publisher = {Copernicus GmbH},
  doi = {10.5194/isprsannals-I-3-293-2012},
  urldate = {2025-09-01},
  langid = {english}
}

@inproceedings{thomasKPConvFlexibleDeformable2019,
  title = {{{KPConv}}: {{Flexible}} and {{Deformable Convolution}} for {{Point Clouds}}},
  booktitle = {2019 {{IEEECVF Int}}. {{Conf}}. {{Comput}}. {{Vis}}. {{ICCV}}},
  author = {Thomas, Hugues and Qi, Charles R. and Deschaud, Jean-Emmanuel and Marcotegui, Beatriz and Goulette, Fran{\c c}ois and Guibas, Leonidas},
  year = 2019,
  pages = {6410--6419},
  doi = {10.1109/ICCV.2019.00651},
  urldate = {2025-09-01}
}

@misc{varneyDALESLargescaleAerial2020,
  title = {{{DALES}}: {{A Large-scale Aerial LiDAR Data Set}} for {{Semantic Segmentation}}},
  author = {Varney, Nina and Asari, Vijayan K. and Graehling, Quinn},
  year = 2020,
  number = {arXiv:2004.11985},
  eprint = {2004.11985},
  primaryclass = {cs},
  publisher = {arXiv},
  doi = {10.48550/arXiv.2004.11985},
  urldate = {2025-09-01},
  archiveprefix = {arXiv}
}

@article{wangDynamicGraphCNN2019,
  title = {Dynamic {{Graph CNN}} for {{Learning}} on {{Point Clouds}}},
  author = {Wang, Yue and Sun, Yongbin and Liu, Ziwei and Sarma, Sanjay E. and Bronstein, Michael M. and Solomon, Justin M.},
  year = 2019,
  journal = {ACM Trans. Graph.},
  volume = {38},
  number = {5},
  pages = {146:1--146:12},
  doi = {10.1145/3326362},
  urldate = {2025-09-01}
}

@article{wangLiDARPointClouds2018,
  title = {{{LiDAR Point Clouds}} to 3-{{D Urban Models}}: {{A Review}}},
  author = {Wang, Ruisheng and Peethambaran, Jiju and Chen, Dong},
  year = 2018,
  journal = {IEEE J. Sel. Top. Appl. Earth Obs. Remote Sens.},
  volume = {11},
  number = {2},
  pages = {606--627},
  doi = {10.1109/JSTARS.2017.2781132},
  urldate = {2025-11-26}
}

@article{wehrAirborneLaserScanning1999,
  title = {Airborne Laser Scanning---an Introduction and Overview},
  author = {Wehr, Aloysius and Lohr, Uwe},
  year = 1999,
  journal = {ISPRS Journal of Photogrammetry and Remote Sensing},
  volume = {54},
  number = {2},
  pages = {68--82},
  doi = {10.1016/S0924-2716(99)00011-8},
  urldate = {2025-11-13}
}

@article{wulderLidarSamplingLargearea2012,
  title = {Lidar Sampling for Large-Area Forest Characterization: {{A}} Review},
  author = {Wulder, Michael A. and White, Joanne C. and Nelson, Ross F. and N{\ae}sset, Erik and {\O}rka, Hans Ole and Coops, Nicholas C. and Hilker, Thomas and Bater, Christopher W. and Gobakken, Terje},
  year = 2012,
  journal = {Remote Sensing of Environment},
  volume = {121},
  pages = {196--209},
  doi = {10.1016/j.rse.2012.02.001},
  urldate = {2025-11-26}
}

@inproceedings{wuPointConvDeepConvolutional2019,
  title = {{{PointConv}}: {{Deep Convolutional Networks}} on {{3D Point Clouds}}},
  booktitle = {2019 {{IEEECVF Conf}}. {{Comput}}. {{Vis}}. {{Pattern Recognit}}. {{CVPR}}},
  author = {Wu, Wenxuan and Qi, Zhongang and Fuxin, Li},
  year = 2019,
  pages = {9613--9622},
  doi = {10.1109/CVPR.2019.00985},
  urldate = {2025-09-01}
}

@article{wuPointTransformerV22022,
  title = {Point {{Transformer V2}}: {{Grouped Vector Attention}} and {{Partition-based Pooling}}},
  author = {Wu, Xiaoyang and Lao, Yixing and Jiang, Li and Liu, Xihui and Zhao, Hengshuang},
  year = 2022,
  journal = {Adv. Neural Inf. Process. Syst.},
  volume = {35},
  pages = {33330--33342},
  doi = {10.48550/arXiv.2210.05666},
  urldate = {2025-09-01},
  langid = {english}
}

@inproceedings{wuPointTransformerV32024,
  title = {Point {{Transformer V3}}: {{Simpler}}, {{Faster}}, {{Stronger}}},
  booktitle = {2024 {{IEEECVF Conf}}. {{Comput}}. {{Vis}}. {{Pattern Recognit}}. {{CVPR}}},
  author = {Wu, Xiaoyang and Jiang, Li and Wang, Peng-Shuai and Liu, Zhijian and Liu, Xihui and Qiao, Yu and Ouyang, Wanli and He, Tong and Zhao, Hengshuang},
  year = 2024,
  pages = {4840--4851},
  doi = {10.1109/CVPR52733.2024.00463},
  urldate = {2025-09-01}
}

@article{xuAdvancedMultiSensorOptical2019,
  title = {Advanced {{Multi-Sensor Optical Remote Sensing}} for {{Urban Land Use}} and {{Land Cover Classification}}: {{Outcome}} of the 2018 {{IEEE GRSS Data Fusion Contest}}},
  author = {Xu, Yonghao and Du, Bo and Zhang, Liangpei and Cerra, Daniele and Pato, Miguel and Carmona, Emiliano and Prasad, Saurabh and Yokoya, Naoto and H{\"a}nsch, Ronny and Le Saux, Bertrand},
  year = 2019,
  journal = {IEEE J. Sel. Top. Appl. Earth Obs. Remote Sens.},
  volume = {12},
  number = {6},
  pages = {1709--1724},
  doi = {10.1109/JSTARS.2019.2911113},
  urldate = {2025-09-01}
}

@article{yanUrbanLandCover2015,
  title = {Urban Land Cover Classification Using Airborne {{LiDAR}} Data: {{A}} Review},
  author = {Yan, Wai Yeung and Shaker, Ahmed and {El-Ashmawy}, Nagwa},
  year = 2015,
  journal = {Remote Sensing of Environment},
  volume = {158},
  pages = {295--310},
  doi = {10.1016/j.rse.2014.11.001},
  urldate = {2025-11-26}
}

@article{yeLASDULargeScaleAerial2020,
  title = {{{LASDU}}: {{A Large-Scale Aerial LiDAR Dataset}} for {{Semantic Labeling}} in {{Dense Urban Areas}}},
  author = {Ye, Zhen and Xu, Yusheng and Huang, Rong and Tong, Xiaohua and Li, Xin and Liu, Xiangfeng and Luan, Kuifeng and Hoegner, Ludwig and Stilla, Uwe},
  year = 2020,
  journal = {ISPRS Int. J. Geo-Inf.},
  volume = {9},
  number = {7},
  pages = {450},
  publisher = {Multidisciplinary Digital Publishing Institute},
  doi = {10.3390/ijgi9070450},
  urldate = {2025-09-01},
  copyright = {http://creativecommons.org/licenses/by/3.0/},
  langid = {english}
}

@article{yooEyeNetMultiscaleMultidensity2025,
  title = {{{EyeNet}}++: {{A Multiscale}} and {{Multidensity Approach}} for {{Outdoor}} 3-{{D Semantic Segmentation Inspired}} by the {{Human Visual Field}}},
  author = {Yoo, Sunghwan and Jeong, Yeonjeong and Sheikholeslami, Mohammad Moein and Sohn, Gunho},
  year = 2025,
  journal = {IEEE Trans. Geosci. Remote Sens.},
  volume = {63},
  pages = {1--19},
  doi = {10.1109/TGRS.2025.3589287},
  urldate = {2025-09-01}
}

@article{yooYUTOSEMANTICLARGE2023,
  title = {{{YUTO SEMANTIC}}: {{A LARGE SCALE AERIAL LIDAR DATASET FOR SEMANTIC SEGMENTATION}}},
  author = {Yoo, S. and Ko, C. and Sohn, G. and Lee, H.},
  year = 2023,
  journal = {Int. Arch. Photogramm. Remote Sens. Spat. Inf. Sci.},
  volume = {XLVIII-1-W2-2023},
  pages = {209--215},
  publisher = {Copernicus GmbH},
  doi = {10.5194/isprs-archives-XLVIII-1-W2-2023-209-2023},
  urldate = {2025-09-01},
  langid = {english}
}

@article{zacharCENAGISALSBENCHMARKNEW2023,
  title = {{{CENAGIS-ALS BENCHMARK}} - {{NEW PROPOSAL FOR DENSE ALS BENCHMARK BASED ON THE REVIEW OF DATASETS AND BENCHMARKS FOR 3D POINT CLOUD SEGMENTATION}}},
  author = {Zachar, P. and Baku{\l}a, K. and Ostrowski, W.},
  year = 2023,
  journal = {Int. Arch. Photogramm. Remote Sens. Spat. Inf. Sci.},
  volume = {XLVIII-1-W3-2023},
  pages = {227--234},
  publisher = {Copernicus GmbH},
  doi = {10.5194/isprs-archives-XLVIII-1-W3-2023-227-2023},
  urldate = {2025-09-01},
  langid = {english}
}

@inproceedings{zhaoPointTransformer2021,
  title = {Point {{Transformer}}},
  booktitle = {2021 {{IEEECVF Int}}. {{Conf}}. {{Comput}}. {{Vis}}. {{ICCV}}},
  author = {Zhao, Hengshuang and Jiang, Li and Jia, Jiaya and Torr, Philip and Koltun, Vladlen},
  year = 2021,
  pages = {16239--16248},
  doi = {10.1109/ICCV48922.2021.01595},
  urldate = {2025-09-01}
}

@misc{zolanvariDublinCityAnnotatedLiDAR2019,
  title = {{{DublinCity}}: {{Annotated LiDAR Point Cloud}} and Its {{Applications}}},
  author = {Zolanvari, S. M. Iman and Ruano, Susana and Rana, Aakanksha and Cummins, Alan and da Silva, Rogerio Eduardo and Rahbar, Morteza and Smolic, Aljosa},
  year = 2019,
  number = {arXiv:1909.03613},
  eprint = {1909.03613},
  primaryclass = {cs},
  publisher = {arXiv},
  doi = {10.48550/arXiv.1909.03613},
  urldate = {2025-09-01},
  archiveprefix = {arXiv}
}
\end{document}